# Parameter Sharing Decoder Pair for Auto Composing

Xu Zhao

**Abstract.** Auto Composing is an active and appealing research area in the past few years, and lots of efforts have been put into inventing more robust models to solve this problem. With the fast evolution of deep learning techniques, some deep neural network-based language models are becoming dominant. Notably, the transformer structure has been proven to be very efficient and promising in modeling texts. However, the transformer-based language models usually contain huge number of parameters and the size of the model is usually too large to put in production for some storage limited applications. In this paper, we propose a parameter sharing decoder pair (PSDP), which reduces the number of parameters dramatically and at the same time maintains the capability of generating understandable and reasonable compositions. Works created by the proposed model are presented to demonstrate the effectiveness of the model.

## 1 INTRODUCTION

Making a computer mimic a human to generate texts is an old challenge and is becoming more and more appealing and active in the past few years, with many novel methods having been developed to deal with this problem. These methods range from traditional Statistic machine translation-based models [1] to neural network-based models [2,3,4]. Some of them have made convincing and promising results and highly inspire researchers to develop more advanced language models to address this challenging problem.

Compared to other methods, the successful adoption of deep learning in so many different areas has led to more and more interest and efforts being put towards using deep neural networks to design models to deal with this tough composing task. These include the recurrent neural networks like LSTM [5], GRU [6], etc. to more recently invented transformer-based language models.

OpenAI released its pre-trained language model called GPT (Generative Pre-training Transformer) [7] in 2018, which first introduced Transformer architecture into the design of a language model and was a big success. Later they released an updated stronger version of the model called GPT2 [8] in 2019 with a few good examples, which further demonstrated the effectiveness of transformer architecture in language model designing. However, pre-training language models usually contain a huge number of parameters and the size of the model is usually too large to put in production for some storage limited applications.

In this paper, we are trying to modify the transformer-based architecture and design a relatively lightweight model, which contains much fewer parameters compared to the classical transformer-based language model, and maintains the capability of generating meaningful, diversified and reasonable texts.

The rest of the paper is organized as follows. In Section 2, the classical transformer based language model is reviewed, then in Section 3, we elaborate on how we modify it and give the design of the proposed model. Section 4 presents and discusses the experiment results. Finally, in Section 5, the conclusion and future work are given.

## 2 REVIEW OF TRANSFORMER-BASED LANGUAGE MODEL

Transformer based language models, like OpenAI's GPT [7], usually only use the transformer decoder. More accurately, they use a modified decoder [9], which contains only the masked multi-head self-attention in each layer. Therefore, we will only give a brief review of the classical transformer decoder and a comparison between the modified decoder and the classical decoder.

A classical transformer decoder usually contains several layers; different layers share the same structure but have different parameters. For a given layer, two key parts are the multi-head attention block and the masked multi-head attention block. The masked multi-head attention gets the inputs from the embedding of the tokens of a raw sentence with additional position information and the multi-head attention gets the inputs from both the output of the encoder and the output of the masked multi-head attention (after dropout and norm).

In the modified decoder, it chops off the multi-head attention and accordingly uses only the masked multi-head attention instead. A simple visual comparison between the classical transformer decoder and the modified transformer decoder is given in Fig 1.

Compared to the classical decoder, the modified decoder contains fewer parameters since in each layer it cuts off one attention block. However, since each layer has its own parameters, the total parameters are still a lot. A simple calculation shows that with the configuration in table 1 the entire 12 layers of a modified decoder contain around 85 million parameters.

**Table 1.** Basic configuration of a modified decoder with 12 layers

| | |
|---|---|
| word embedding size | 768 |
| num of attention heads | 12 |
| size per head | 64 |
| hidden size | 768 |
| feedforward network size | 3072 |
| num of layers | 12 |

In the next section, we'll introduce the parameter sharing decoder pair (PSDP) as a way to reduce model parameters.

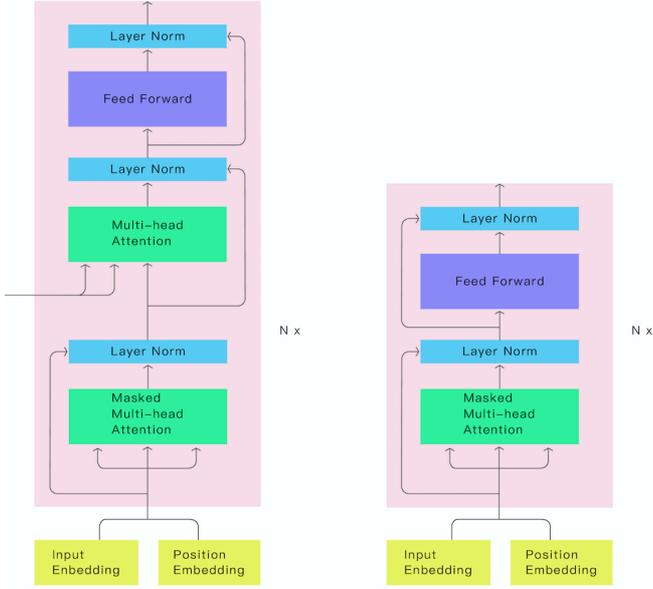

**Figure 1.** A comparison between classical decoder and modified decoder. (Left) classical decoder; (Right) Modified Decoder. N usually equals to 12 for basic configuration.

## 3   PSDP: Parameter Sharing Decoder Pair

From the above section, we can see that a modified decoder with basic configurations usually has a huge number of parameters to learn. It means not only do we need to spend more effort to learn the parameters well, but the generated model usually takes a lot of space to store, which makes it a big challenge to some applications where only limited storage is available.

In order to address this issue to some extent, we introduce a new design which uses parameter sharing decoder pair (PSDP) to model the data. Concretely, in PSDP we have two smaller modified transformer decoders, each of which has its own set of parameters for both the masked multi-head attention and the feedforward network, with the parameters being shared across all the layers within the decoders. The two smaller decoders are tied together by concatenating their outputs and then another mapping is applied to down-project the combined outputs back to the embedding size. Standard layer normalization as used in each of the attention layers is applied before the final output. Note that we add the average of the outputs of the two decoders in the last layer normalization. Also, we remove all the dropouts inside the model. The inspiration comes from [10], where in the paper the author points out that adding dropout potentially hurts the performance of the transformer-based models.

Fig 2 provides a visual illustration of the proposed architecture of PSDP; the parts that need to share parameters are shadowed. Masked multi-head attention and feedforward network of the two smaller decoders are assigned different colors (green and pink) to indicate they have different sets of parameters.

Obviously in PSDP if we set N equal to 6, we still have 12 layers in total, but since we are doing parameters sharing, the total number of parameters decreases a lot. By a rough calculation, we can see that by using the same configuration as in table 1, the entire 12 layers of PSDP contains around 14.2 million parameters which is about 17% of the original decoder (85 million). However, with the same configuration PSDP does introduce about 1.18 million additional parameters in the mapping step after concatenating the two outputs, but the overall parameters reduction is still considerable.

The optimization objective is to minimize the cross entropy between the expected outputs and the actual outputs of PSDP. We are using Adam optimizer to learn the parameters of the model.

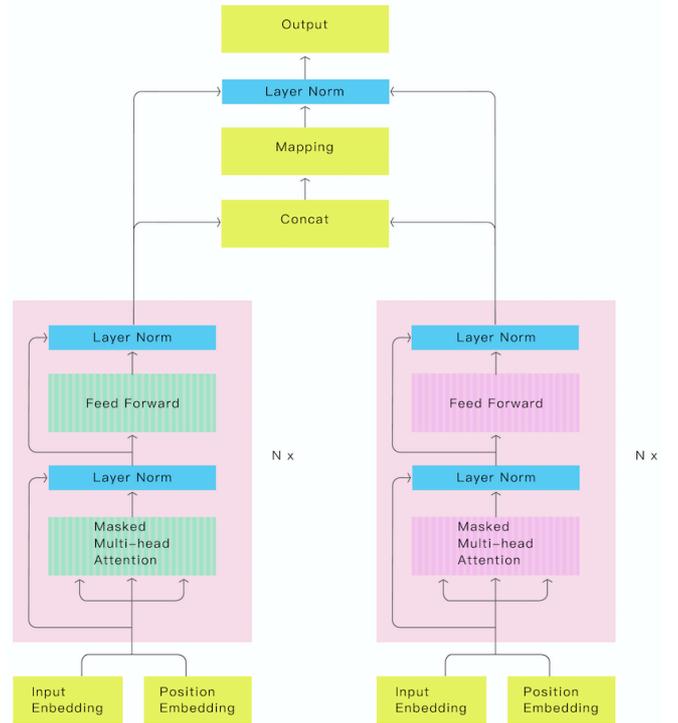

**Figure 2.** A visual illustration of the proposed Parameter Sharing Decoder Pair (PSDP)

## 4   Experiments and Discussion

In this section, we present the experiment results of using PSDP to generate compositions. In all the following experiments, we are using WordPiece tokenizer [11] to tokenize the texts. The compositions are generated token by token and at each time point the top k candidates with the highest possibility are selected. We then randomly sample one token based on their possibility. We set k to be 10 in English tasks and 15 in Chinese tasks. All the training processes are executed in a Tesla P100-SXM2 GPU with 16G memory.

## 4.1 English compositions completion

The training data used in this model is the Wikipedia articles with minimal filtering, like ignoring sentences less than 10 words and removing html tags. We use two 12 layers modified decoders. The training step is set to be 500,000 and batch size is 8. Here, we use the following configuration as in Table 2:

**Table 2.** Configuration of PSDP for English compositions generation

| | |
|---|---|
| word embedding size | 1200 |
| num of attention heads | 12 |
| size per head | 100 |
| hidden size | 1200 |
| feedforward network size | 4800 |
| num of layers (each decoder) | 12 |
| Maximum sequence length | 128 |

The trained model is about 300 MB. (The model will be about 1730M MB if we stack the decoders without sharing parameters.) The training takes around 70 hours. (the model still underfits the data)

Here we showcase several generated compositions by PSDP. For each of these compositions, we first give it a start prompt, and let PSDP finish the whole story. We are trying to make the generated stories be around 100 tokens and we are using auto-regressive to generate each token. All the characters are transformed to lowercase.

**Example 1:**
**Start prompt:** many species live in the jungle
**Compositions generated by PSDP:**
many species live in the jungle , and they can be found growing from near - surface to high – altitude . these may include small mammals , birds such as hawks , eagles , kites , herons , and lesser owls . some of these are very rare or endangered species that are threatened by habitat loss . this makes up about 20 % of the population ( around 40 ) who live in forest and have a natural range of habitats . there have been several attempts to save the area since the late 19th century .

**Example 2:**
**Start prompt:** the scientists finally found the cave
**Compositions generated by PSDP:**
the scientists finally found the cave , and discovered it as a result of the collapse of the ice sheets on earth . they then began to search for the remains . the first known discovery came from a strange hole near the bottom floor of the cave . this finding was discovered by the british geologist , john smith , who described the caves as " a great part of the earth ' s history " . the second discovery was by the australian geological survey ( australia ) , which had concluded that the rock was not a significant part of the cave itself . however , this cave became the site of one of the world ' s most famous discoveries .

**Example 3:**
**Start prompt:** after graduated from
**Compositions generated by PSDP:**
after graduated from the university of california , berkeley , she began writing for the " los angeles times " . her first novel was published in 1977 , and received numerous honors . " the new york times " described her as " one of the most influential feminist fiction writers in the united states " . her novels have been translated into several languages , including spanish , french , english , japanese , portuguese , and italian . she has appeared in many anthologies since the early 1990s . among these are " el amor " ( 1995 ) , the anthology " el lazar " ( 2004 ) , and " les ques " ( 2007 ) .

From the above works, we can see PSDP can complete the story with understandable and reasonable sentences. However, due to our random sampling strategy we may bring in some uncertainties, which may cause the generated compositions to look weird. We do observe the following phenomena:

1. In some cases, the generated compositions are not in line with facts. Like in example 3, it says an American writer's novels are translated into English. Another example is as follows:
   "deep ocean fishes are mysterious creatures that live in the sea, including the giant "big horn" and various large species of birds such as those from north america."
   Obviously, birds are not fish and cannot live in the deep ocean.
2. Sometimes it may generate repeat chunks like the following paragraph:
   "the most popular foods available at the factory include chicken soup, chicken soup, and beef meat."
3. In some cases, two consecutive sentences may not be coherent.

Overall, it may take several tries to complete one high quality paragraph. One trick used here is if a token has been generated before we will try it again for up to 3 times. Another trick is to let it generate the whole story chunk by chunk (chunk here can be a sentence or any consecutive tokens), each time we pick the best next chunk based on the current context and then add this best chunk back to the context to generate the next chunk.

## 4.2 Chinese couplet completion

In order to further test the composing capability of PSDP, we attempt the Chinese Couplet task. Chinese Couplets have important cultural heritage with a long history in China. A classical couplet usually contains two sentences with the same length. The characters on each line should correspond to each other and furthermore each sentence should follow a special tone pattern.

In our experiments, unlike two-steps methods [12], which first pre-train a language model and then do fine-tuning, our model is directly trained on the Couplet dataset. The training data is collected from a public dataset[1], which contains around 770,000 couplets in total. We use two 12 layers modified decoders. All the other configurations are listed in table 3. The training step is set to be 250,000 and the batch size is 8. The entire training takes about 19 hours. The trained model is about 254 MB. (The model will be about 1680 MB if we stack the decoders without sharing parameters.)

---
[1] https://github.com/wb14123/couplet-dataset.

Table 3. Configuration of PSDP for Chinese couplets generation

| | |
|---|---|
| word embedding size | 1200 |
| num of attention heads | 12 |
| size per head | 100 |
| hidden size | 1200 |
| feedforward network size | 4800 |
| num of layers (each decoder) | 12 |
| Maximum sequence length | 64 |

To test the performance, we give the model the first sentence of a couplet and let the model generate the corresponding second sentence. Several couplets generated by PSDP are shown in table 4.

From the results we can see that due to the random sampling strategy the model can generate diversified second sentences for a fixed first sentence, and they parallel with each other pretty neatly.

However, sometimes it will generate some unknown words, which makes the couplets look strange, but most of the cases it works well. We will try to introduce a pre-trained Chinese language model by using PSDP and then do fine-tuning on top of it to see if we can enhance the performance.

Table 4. Couplets generated by PSDP.

| 上联 (the first sentence of the couplet) | 下联 (the second sentence of the couplet) |
|---|---|
| 雨后清风拂嫩草。(Breeze after rain gently caresses tender grass.) | 云中紫燕剪斜阳。(Lilac swallows in the clouds cut through the setting sun.) |
| | 花前细雨润新苗。(Drizzle in front of the flowers moisten sprouts.) |
| | 江中皓月照归舟。(Shining moon over the river illuminates the returning boats.) |
| 云罩山间路。(Clouds blanket the mountain path.) | 月勾柳下舟。(Moon hooks the boat under the willow.) |
| | 日沉水底天。(Sun's reflection floats in the blue beneath the water.) |
| | 雾封岭上松。(Mist seals the pines on the ridge.) |
| 十里桃花飘香远。(The scent of peach blossoms wafts out far.) | 一轮红日破雾开。(Red round sun spikes through the fog.) |
| | 千年古木落叶青。( Ancient tree sheds green leaves. ) |
| | 一湾碧水映日辉。(Bay of jade water reflects the glow of the sun.) |

## 5 CONCLUSIONS AND FUTURE WORK

In this paper, we proposed a parameter sharing decoder pair (PSDP) for auto composing. Unlike stacking all the attention layers together, we use two smaller decoders and share parameters across all the layers. The generated model employs much fewer parameters and maintains the capability to compose understandable and reasonable compositions at the same time. Experiments on English compositions completion and Chinese Couplet composing are conducted to demonstrate the effectiveness of the proposed method.

Future studies will aim to take advantage of the design to introduce parallel inferences, which may cut the inference time dramatically. Also, we are going to build a hierarchical decoder array to explore the limit of the model as well. Another improvement will be trying to increase the size of the training data like WebText used in OpenAI GPT2.